\documentclass[conference]{IEEEtran}
\makeatletter
\def\ps@headings{%
\def\@oddhead{\mbox{}\scriptsize\rightmark \hfil \thepage}%
\def\@evenhead{\scriptsize\thepage \hfil \leftmark\mbox{}}%
\def\@oddfoot{}%
\def\@evenfoot{}}
\makeatother
\usepackage{amsmath,amssymb,amsfonts}
\usepackage{mathtools} 
\usepackage{commath}
\usepackage{textcomp}
\usepackage{comment}
\usepackage{xcolor}
\usepackage{graphicx}
\usepackage{caption}
\usepackage{subcaption}
\usepackage{xfrac}
\usepackage{upgreek}
\usepackage{algorithm}
\usepackage{algorithmic}
\usepackage{makecell}
\usepackage{enumitem}
\usepackage{multirow}
\usepackage{color, colortbl}
\usepackage{makecell}
\usepackage{pifont}
\usepackage{soul}
\usepackage{makecell}
\usepackage{framed}
\usepackage{tikz}

\definecolor{Gray}{gray}{0.8}
\definecolor{Dark_Gray}{gray}{.6}

\def\3rd{$3^{rd}$ party app}

\title{Deep Learning Approach Protecting Privacy in Camera-Based Critical Applications\\
{\footnotesize}
\thanks{}
}

\author{\IEEEauthorblockN{Gautham Ramajayam}
\IEEEauthorblockA{\textit{\small Stony Brook University} \\
\small gramajayam@cs.stonybrook.edu}
\and
\IEEEauthorblockN{Tao Sun}
\IEEEauthorblockA{\textit{\small Stony Brook University} \\
\small tao@cs.stonybrook.edu}
\and
\IEEEauthorblockN{Chiu C. Tan}
\IEEEauthorblockA{\textit{\small Temple University} \\
\small cctan@temple.edu}
\and
\IEEEauthorblockN{Lannan Luo}
\IEEEauthorblockA{\textit{\small University of South Carolina} \\
\small lluo@cse.sc.edu}
\and
\IEEEauthorblockN{Haibin Ling}
\IEEEauthorblockA{\textit{\small Stony Brook University} \\
\small hling@cs.stonybrook.edu}
}

\begin{document}
\maketitle

\begin{abstract}
Many critical applications rely on cameras to capture video footage for analytical purposes. This has led to concerns about these cameras accidentally capturing more information than is necessary. In this paper, we propose a deep learning approach towards protecting privacy in camera-based systems. Instead of specifying specific objects (e.g. faces) are privacy sensitive, our technique distinguishes between salient (visually prominent) and non-salient objects based on the intuition that the latter is  unlikely to be needed by the application.       

\end{abstract}


\section{Introduction}
Many of today's critical applications rely on cameras. Examples include applications used for public safety~\cite{koh2016improve}, surveillance~\cite{sehrawat2017surveillance}, health care~\cite{chen2014camera}. manufacturing~\cite{bottani2019augmented}, and so on. The common feature of all such systems is a camera that continuously recording the user and his environment, with the video feed being transmitted to the backend servers for processing. 

Since the camera's indiscriminate recording of its surroundings will naturally capture information unrelated to the system's intent, it is unsurprising that privacy has emerged as a major concern. A major research direction in addressing this problem is the use of \textit{obfuscation} to protect privacy~\cite{padilla2015visual}. This is where specific objects are (e.g. faces, license plates, computer screens, etc.) are first identified, and then distorted, so as to make them unintelligible (e.g. blurring the license plate) to the viewer of the video recordings. In this paper, we will use the general term ``blurring" to refer to this distortion process, though in practice, other methods such as blacked outs can also be used.  

The typical process of obfuscation is as following. The system developer will first identify a list of objects that are privacy sensitive (e.g. faces, computer screens, and so on) and then design computer vision algorithms to automatically identify these objects in the video and then blur then out. End users may be able to configure the system to selectively blur certain objects (e.g. blur logos but not license plates), depending on the application requirements. 

The limitation with the current approach of obfuscation is that the concept of privacy cannot be easily reduced to a set of objects to blur. The list of privacy sensitive objects as determined by the system developer, may not necessarily match different users' requirements. Having too many options will overburden the user configuration process, while having too few options may result in the user being unable to adequately protect privacy. Furthermore, whether an object reveals privacy or not, is dependent on the \textbf{context} of the situation. For example, whether a person's face needs to be blurred to protect privacy depends on a specific context. Someone in the background that is taking to the camera is probably having a conversation and should not be blurred, while the same person that turns around and talks to someone else, probably should be blurred. 

Our approach to this problem is based on deep learning. Different from previous methods that detect a list of predefined privacy-sensitive objects, our method only detect a single class of privacy regions. We utilize a key concept called \emph{Visual Saliency} \cite{WangLFSLY21pami}, meaning visually distinctive objects or regions in an image, with a prior knowledge that privacy regions are less likely to be salient and vice versa. This mutually exclusive relationship makes it possible to detect privacy-sensitive objects while considering their specific context. We first formulate a two-stage method that refines results from a privacy object detector using saliency scores. Then we propose a hybrid model that combines object detector and saliency detector together and can be trained in an end-to-end way. 

To validate the proposed methods, we collected a dataset of video call scenario and annotated the privacy sensitive objects in the images of the dataset. Then, the proposed methods are tested on the dataset along with standard object detection algorithms. The effectiveness of the proposed approach is clearly validated by the experimental results.

\section{Related Work}
 
Camera privacy is a critical issue in computer vision as camera-captured images contain extensive visual details that can lead to privacy leakage~\cite{tonge2020image,orekondy2017towards,yu2016iprivacy,orekondy2018connecting,tonge2019privacy,yu2018leveraging,shen2019human,yang2013appintent,zerr2012picalert,Lang&Ling15tip,kummerer2016deepgaze}. Several works have explored detecting privacy attributes from images such as license plates detection, age estimation from facial photographs, social relationships mining, face detection, landmark detection and occupation recognition. Another line of research aims to protect images from malicious recognition by adding adversarial perturbation to the images~\cite{oh2017adversarial}. The perturbation is non-perceivable to human but destructive to visual models.

Object detection is one of the fundamental tasks in computer vision. The goal is to recognize and locate predefined objects in an image. There are two major categories of deep learning based object detection methods, one-stage methods and two-stage methods. YOLO~\cite{redmon2016you} is one of the most representative one-stage methods. It frames object detection as a regression problem, and predicts the bounding boxes and class probabilities directly from full images without post-processing. YOLO is extremely fast due to its unified architecture. YOLOv3~\cite{redmon2018yolov3} improves over YOLO by predicting across 3 different scales. This makes it more accurate to detect objects of different sizes. Faster R-CNN~\cite{ren2015faster}, different from YOLO, needs to generate potential bounding boxes first and then classify these proposed boxes. The results are refined with post-processing. Faster R-CNN represents two-stage methods that are accurate but slow and hard to optimize.

Saliency detection has different motivation compared to generic object detection. Salient object detection aims to find most visually distinctive objects or regions in an image. Whether one object is salient depends on its context. For example, the face of a person could be a salient one when he/she is talking to another one in a video conference, but not when the video is in presentation mode. Saliency detection can be implemented using deep neural networks with pixel-level classification losses. A recent survey can be found in~\cite{WangLFSLY21pami}. Among many salient object or saliency detection algorithms, two of them are mostly related to our study: Hou et al.~\cite{hou2017deeply} builds upon VGGNet, and fuses classification losses at 6 different scales. Qin et al.~\cite{qin2020u2} proposes a new architecture named U$^2$-Net that does not rely on pretrained backbones. It adopts a two-level nested U-structure and novel Residual U-blocks. Saliency detection has an inborn connection with privacy detection. It is less likely that a salient region contains privacy-sensitive objects. 
\section{Using Deep Learning to Protect Privacy}


One key step in protecting privacy is to detect privacy regions in a camera-captured image. This privacy detection, or more generally image detection, is one of the fundamental tasks in computer vision. Deep learning based methods have greatly improved the detection accuracy on benchmarks like MS-COCO~\cite{lin2014microsoft}. Besides their good performance, a significant advantage over conventional computer vision methods is that they can be easily transferred to new datasets. For example, we can take a YOLOv3 model for generic objection detection pretrained on MSCOCO and finetuned it on the privacy protection images for privacy detection. The powerful representation ability of deep learning enables context-aware privacy detection where the the spatial information of surroundings to the persons and temporal information of current activities are considered. 

Usually, saliency regions in an image are less likely to be privacy sensitive. With this prior knowledge, we can use a saliency detection model to refine the results from privacy detector. Depending on the way to use this information, we investigate two types of privacy detection approaches: a two-stage method and an end-to-end one.



\subsection{Two-stage method}
In the method, the privacy region detector and saliency detector are trained independently. To make a prediction on an image, the privacy detector first detect candidate privacy regions, then these candidates are filtered based on saliency detection results. For the first stage, we tried two representative image detectors, YOLOv3 and Faster-RCNN, as the privacy region detector, and chose the \textit{Deeply Supervised Salient Object Detection with Short Connections} as described in the related work section as the saliency detector. The detailed procedure of saliency filtering are listed in the evaluation part.

\subsection{End-to-end method}
\begin{figure}[h]
\centerline{\includegraphics[trim=2cm 1cm 3cm 1.5cm,clip=true,width=1\linewidth]{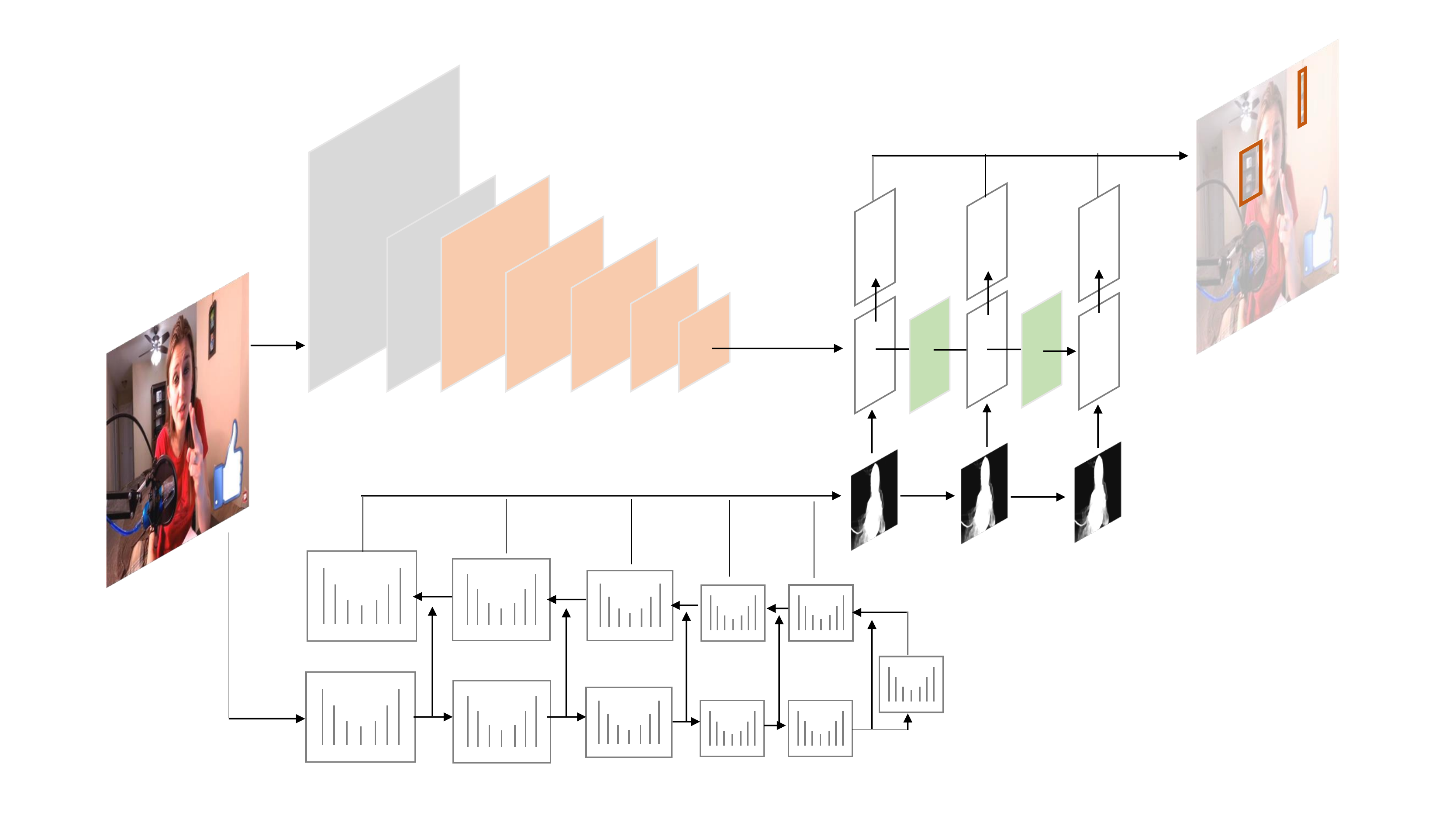}}
\caption{Architecture of hybrid YOLOv3+$U^2$-Net. The lower subnet is the $U^2$-Net for saliency detection and the upper subnet is the YOLOv3 for privacy object detection. Residual connections and module meanings are omitted for clarity. Please refer to original paper for details.}
\label{fig:e2e_method}
\end{figure}

A potential drawback of previous the two stage method is that the generation of privacy regions does not consider the saliency information, which can be sub-optimal. Rather than manually thresholding with saliency map, an alternative way is to take the saliency map as additional features to the privacy region detector and let model automatically learn the relationship between privacy region and saliency region. Figure \ref{fig:e2e_method} shows the architecture of the end-to-end method using YOLOv3 as privacy region detector and $U^2$-Net as saliency detector. YOLOv3 and U$^2$-Net takes the same original image as the network input. Then the output saliency map from U$^2$-Net are viewed as additional feature presentations and added to the latent feature of YOLOv3 before generating privacy regions. Since YOLOv3 works at three different resolutions, the saliency map is rescaled to fit the corresponding ones. In this hybrid model, the generation of privacy regions not only relies on object semantics but also on contextual saliency.

\section{Evaluation}
\subsection{Data collection}
During the COVID-19 pandemic, an increasing number of video calls are made every day in teaching and business activities. There is, however, a great risk of privacy leakage where the camera captures irrelevant background regions that are not supposed to be revealed to the audiences. 

\begin{figure}[h]
\center
\includegraphics[trim=1cm 1cm 1cm 1cm,clip=true,width=0.9\linewidth]{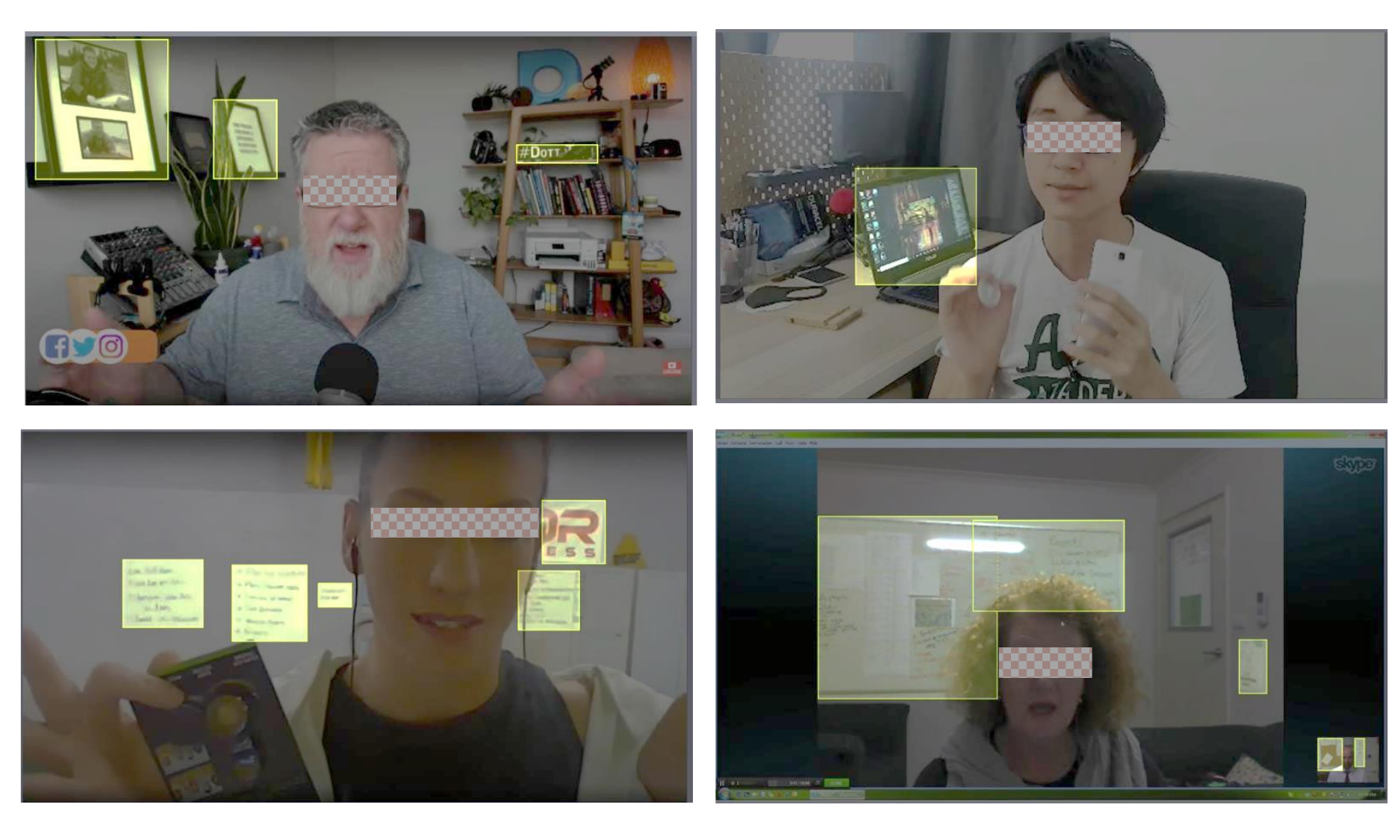}
\caption{Exemplar images with notations from the \textit{YouTube Video Call Dataset}. Eye regions are blocked on purpose for visualization (same for Figure 3 as well). }
\label{fig:example_image}
\end{figure}

To study context-awareness privacy protection in this situation, we collect some YouTube videos that are about video call, and extract a total of 1,000 images from them. In each image, the privacy regions, e.g., containing faces and text depending on the context, are manually labeled. We name this dataset as \textit{YouTube Video Call Dataset}. Figure \ref{fig:example_image} shows exemplar images and their ground-truth notations of privacy regions.

\subsection{Experiment details}
We follow standard evaluation procedure and split the \textit{YouTube Video Call Dataset} by 80\% for training and 20\% for testing. Each model is trained on the training set, and then evaluated on the test set using mean average precision (mAP). This metric measures how well the predicted privacy regions are consistent with the ground-truth privacy regions.

\subsubsection{YOLOv3 with Manual Thresholding (MT) using saliency}
A YOLOv3 model pretrained on MS-COCO~\cite{lin2014microsoft} is fine-tuned on the training images of the collected dataset to detect privacy regions. Meanwhile, a pretrained saliency detector is fine-tuned on the training images to detect salient regions. In Manual Thresholding (MT) using saliency step, the predicted privacy regions from the YOLOv3 model are refined with the saliency map. If the average saliency score of a detected privacy region is above 0.5, it will be rejected and not considered as a privacy region.

\subsubsection{Faster R-CNN with Manual Thresholding using Saliency}
This model replaces the YOLOv3 in previous method with the pretrained Faster R-CNN model. It adopts the same procedure to refine detected privacy regions with saliency score thresholding.

\subsubsection{Hybrid YOLOv3+$U^2$-Net}
The hybrid model combines YOLOv3 and $U^2$-Net into one integrated model and train in an end-to-end way. The saliency map from $U^2$-Net are input as additional feature to the YOLOv3 model. The whole model is trained on the training images for 100 epochs. To further analyze the effect of saliency map, we also add another step of manual threshold as we did in previous methods.

\subsection{Quantitative results}

The evaluation results are listed in Table \ref{tab:results}. As can be seen from the table, using saliency in visual privacy detection clearly improves the mAP compared to not using saliency. YOLOv3 performs comparable to Faster R-CNN, though its mAP is slightly higher. Hybrid YOLOv3+U$^2$-Net improves over YOLOv3 only, showing the effect of combining saliency into privacy detection procedure. An interesting point is that the post-processing of manually thresholding of saliency works better than end-to-end training. The reasons may be that our current dataset is not large enough to implicitly learn the mutual exclusive relationship between privacy region and saliency region. This can be improved by shifting the region classification probability of the privacy detector, rather than using saliency map as latent feature only. We leave this for future work.

\begin{table}[h]
\begin{center}
\caption{Comparison of different privacy detection methods on the test set. Note: MT is short for ``manual threshold".}
 \begin{tabular}{c| c} 
 \hline\hline
 \textbf{Method} & \textbf{mAP (\%)}  \\ [0.5ex] 
 \hline\hline
 YOLOv3  & 27.4    \\
 \hline
 Faster R-CNN & 25.5   \\
 \hline
 YOLOv3 w/ MT  & 35.1    \\
 \hline
 Faster R-CNN w/ MT & 33.6   \\
 \hline
 Hybrid YOLOv3+$U^2$-Net  & 29.8    \\
 \hline
 Hybrid YOLOv3+$U^2$-Net w/ MT & 34.7   \\
 \hline\hline
\end{tabular}\label{tab:results}
\end{center}
\end{table}

\subsection{Qualitative  results}
\begin{figure}
        \centering
        \begin{subfigure}[b]{9cm}
            \centering
            \includegraphics[trim=.5cm 0cm .5cm 0cm,clip=true, width=\textwidth]{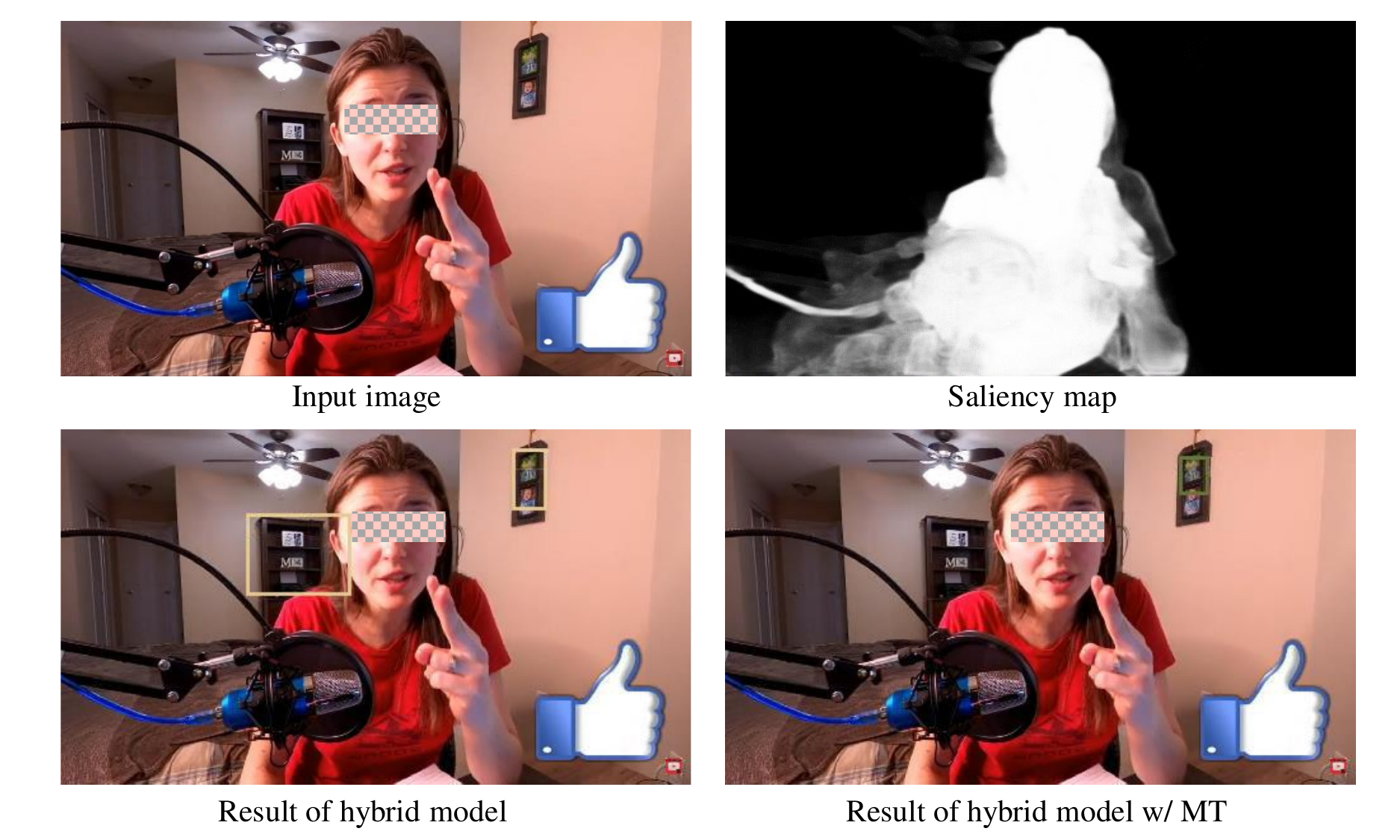}
            \end{subfigure}
        \begin{subfigure}[b]{9cm}  
            \centering 
            \includegraphics[trim=.5cm 0cm .5cm 0cm,clip=true,width=1\linewidth]{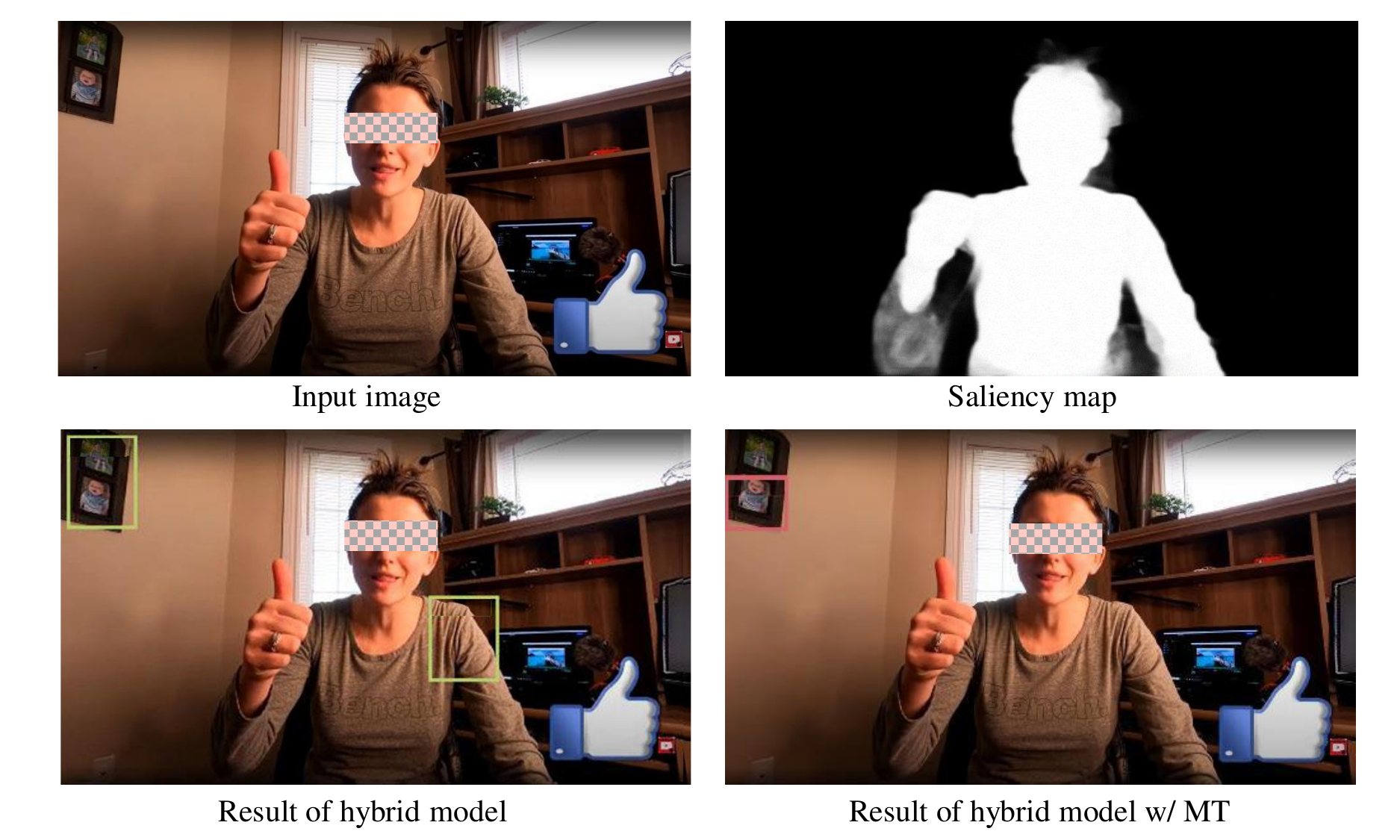}
        \end{subfigure}
    \caption{Privacy region detection results with hybrid model and manual thresholding.}
    \label{fig:qualititive_result}
    \end{figure}

In this section, we provide some qualitative results. Figure \ref{fig:qualititive_result} shows two test images and their privacy region detection results. As can been seen, the saliency map indicates the most prominent objects in the images, i.e., persons. The hybrid model detects privacy regions in the images, but there are false positive results. After manually thresholding with saliency map, these false positive results are removed.

\section{Conclusion and Future Work}

In this paper, we present the deep learning approach to protect privacy in camera-based critical applications. We point out the limitations of current approaches in improper predetermined privacy-sensitive objects and under-exploitation of context of privacy situation. We utilize the mutual exclusive relationship beween privacy regions and saliency regions, and propose a deep learning-based privacy-sensitive object detection approach. Our experiments on the collected video call dataset demonstrate its effectiveness.

Our results are very preliminary and there are several future works to explore. Currently, end-to-end training is not as effective as manually thresholding. It is worthwhile to study how to bring saliency map into the decision procedure of the privacy object detector. For example, the classification score as privacy object may be shifted based on the saliency score. In video data, the temporal information from adjacent frames could be used to better infer current activates. This is critical as whether an object is privacy sensitivity is highly related to the person's intention. Moreover, more effective ways, such as earlier feature fusion, for integrating the saliency information into privacy sensitivity detection can potentially improve the detection accuracy. Finally, another direction is to inpaint privacy-sensitive regions to make it natural rather than simply obfuscating the regions.  

\bibliographystyle{plain}
\bibliography{bibfile}

\end{document}